\documentclass{article}
\usepackage{ijcai24}
\usepackage{times}
\usepackage{soul}
\usepackage{url}
\usepackage[hidelinks]{hyperref}
\usepackage[utf8]{inputenc}
\usepackage[small]{caption}
\usepackage{graphicx}
\usepackage{amsmath}
\usepackage{amsthm}
\usepackage{booktabs}
\usepackage{algorithm}
\usepackage{algorithmic}
\usepackage{amsfonts,amssymb}
\usepackage{dsserif}
\usepackage{multirow}
\usepackage{bbding}
\usepackage[switch]{lineno}

\title{Focus on the Whole Character: Discriminative Character Modeling for Scene Text Recognition}

\author{
    Bangbang Zhou\and
    Yadong Qu\and
    Zixiao Wang\and
    Zicheng Li\and
    Boqiang Zhang\and
    Hongtao Xie\footnote{Corresponding Author}
    \affiliations
    University of Science and Technology of China, Hefei, China
    \emails
    \{bangzhou01, qqqyd, wzx99, lizicheng, cyril\}@mail.ustc.edu.cn,
    htxie@ustc.edu.cn
}

\begin{document}

\maketitle

\begin{abstract}
Recently, scene text recognition (STR) models have shown significant performance improvements.  
However, existing models still encounter difficulties in recognizing challenging texts that involve factors such as
severely distorted and perspective characters. 
These challenging texts mainly cause two problems: (1) Large Intra-Class Variance. (2) Small Inter-Class Variance. 
An extremely distorted character may prominently differ visually from other characters within the same category, while the variance between characters from different classes is relatively small. 
To address the above issues, we propose a novel method that enriches the character features to enhance the discriminability of characters. 
Firstly, we propose the Character-Aware Constraint Encoder (CACE) with multiple blocks stacked. 
CACE introduces a decay matrix in each block to explicitly guide the attention region for each token. 
By continuously employing the decay matrix, CACE enables tokens to perceive morphological information at the character level. 
Secondly, an Intra-Inter Consistency Loss ($\text{I}^2\text{CL}$) is introduced to consider intra-class compactness and inter-class separability at feature space. 
$\text{I}^2\text{CL}$ improves the discriminative capability of features by learning a long-term memory unit for each character category. 
Trained with synthetic data, our model achieves state-of-the-art performance on common benchmarks (94.1\% accuracy) and Union14M-Benchmark (61.6\% accuracy). Code is available at https://github.com/bang123-box/CFE.
\end{abstract}

\begin{figure}[!t]
     \centering
\includegraphics[width=1\linewidth]{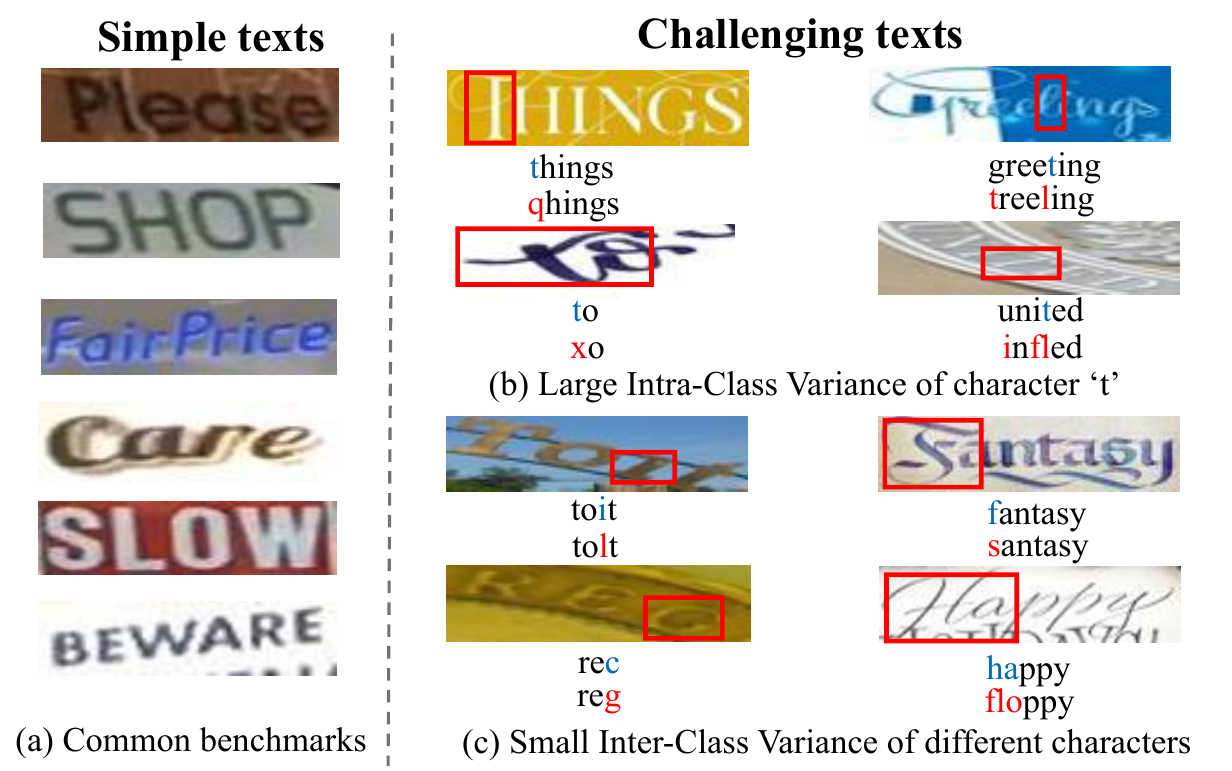}
     \caption{Differences between simple and challenging texts. (a) Simple texts are singular in style and uniform in size. (b) With variations in appearances and size, the character `t’ is misrecognized. (c) The similarity in appearances of different category characters leads to wrong recognition. The first
line is the label and the second line is the prediction with our baseline model. The incorrectly recognized characters are highlighted in red.
     }
\label{fig:motivation}
\end{figure}
\section{Introduction}
Scene Text Recognition (STR) aims to recognize character sequences from cropped text images~\cite{parseq,zhangbo,lister,rmt}. Existing STR methods adeptly read texts encompassing billboards, road signs, checks, {\em etc}. 
However, with societal advancements, the demands on STR models are no longer limited to performing well in simple texts, but also need to improve their performance in challenging texts.

The recent methods have obtained superior performance on six common benchmarks~\cite{CUTE,IIIT,SVT,SVTP,IC13,IC15} which are easy to recognize as shown in Figure \ref{fig:motivation}(a).
However, with the introduction of a challenging Union14M-Benchmark~\cite{union14m}, existing STR models perform poorly on Curve, Artistic, and Contextless datasets.
For example, MGP-Base~\cite{MGP} only achieves the accuracy rates of 55.2\%, 52.8\%, and 48.4\% on them, respectively. 
We believe that the poor performance on challenging texts is primarily due to two ignored issues.

The first issue is summarized as Large Intra-class Variance (LISV). Figure \ref{fig:motivation}(b) illustrates the different instances of the character `t'. There are variations in the visual appearances, shape, and size, which lead to errors in recognition results. 
We attribute the misrecognition of characters to the large variance of the same category characters. 
Due to the existence of LICV, the discriminative features of characters will be weakened, eventually leading to the misrecognition of characters.
To solve LICV, it's essential to enhance the discriminability of features by encoding local patterns or structures within the character. 
This helps model the correlation between the components of a character, capturing discriminative information more comprehensively. 
However, existing STR encoders have difficulty learning the intra-character local patterns.
CNN-based encoders~\cite{CRNN,trba} can learn local features, but the receptive field is too small to perceive the information of the whole character. Transformer-based encoders~\cite{nrtr,parseq} hardly focus on the local information at  character level due to the global modeling during self-attention. 
Thus, improving the ability to encode local patterns ({\em e}.{\em g}., stroke, morphology, {\em etc}.) at character level is a crucial step in solving LICV.

The second problem is Small Inter-Class Variance (SICV). 
As shown in Figure \ref{fig:motivation}(c), our baseline recognizes some characters as other different category characters due to their similar visual appearances. 
For example, the word `happy' is misrecognized as `floppy', which also conforms to linguistic rules.
Therefore, using additional language models or linguistic information does not help solve the SICV problem.
Furthermore, we argue that SICV can lead to a mixed distribution of different category character features in the decoding space (demonstrated formally in Sec. \ref{sec:visualization}). 
To solve SICV and LICV together, CornerTransformer~\cite {wordart} designs a Character Contrastive loss (CC loss) to bring the same category characters closer and separate characters from different classes. 
However, since CC loss only considers the character feature distribution within a batch, it greatly limits the diversity and richness of the global character feature distribution.
Hence, how to utilize global character feature distribution is a great challenge for solving LICV and SICV.

To address the above issues, we propose a Character Features Enriched model (CFE) to obtain the discriminative character features from two aspects. 
Firstly, the Character-Aware Constraint Encoder (CACE) is proposed to perceive the local patterns such as the morphological information. 
For CACE, the discriminative features are extracted by multiple stacked blocks.
In each block, we design a decay matrix $\textbf{D}$ to attenuate the self-attention mechanism according to the spatial distances between visual tokens.
The greater the distances between tokens, the less attention is paid to them. 
Compared to the vanilla self-attention, the block can alleviate the noise interference from other characters and pay more attention to the region near each token. 
In this way, CACE encodes the local patterns and relationships between the components of each character.
Additionally, to fully utilize the visual features output from the multiple blocks, a fusion strategy is employed to merge them. 
Secondly, an Intra-Inter Consistency Loss ($\text{I}^2\text{CL}$) is introduced to solve LICV and SICV.
Based on the contrastive learning~\cite{ccloss,wordart,conclr}, $\text{I}^2\text{CL}$ further predefines a long-term memory unit for each character category.
For each character in a batch, its positive example refers to the unit with the same character category, while the other units serve as negative samples.
In the training, $\text{I}^2\text{CL}$ updates these memory units based on each character. 
Different from~\cite{wordart,conclr} that only consider the local distribution, $\text{I}^2\text{CL}$ can efficiently represent all characters by learning a discrete distribution for long-term memory units.
Finally, all characters will be tightly distributed around the memory units according to their categories. 
This ensures the intra-class compactness and inter-class separability, and improves the discriminability of characters.
Compared with previous methods, we have fewer training parameters, while achieving better performance.

The main contributions of our work are as follows:
\begin{itemize}
\item We point out that the LICV and SICV issues in challenging texts lead to poor performance of the STR models, and propose a novel approach to effectively handle the two issues.
    \item We design a Character-Aware Constraint Encoder to 
    focus on the local patterns of character level, which utilizes the morphological information to enrich features.
    \item We introduce an Intra-Inter Consistency Loss to reduce intra-class variance and increase inter-class variance by learning a set of long-term memory units.
    \item Experiments on common benchmarks and Union14-Benchmark demonstrate that our CFE surpasses state-of-the-art performance, with accuracy rates of 94.1\% and 61.6\%, respectively.
\end{itemize}

\begin{figure*}[!t]
\centering
\includegraphics[width=0.75\textwidth] {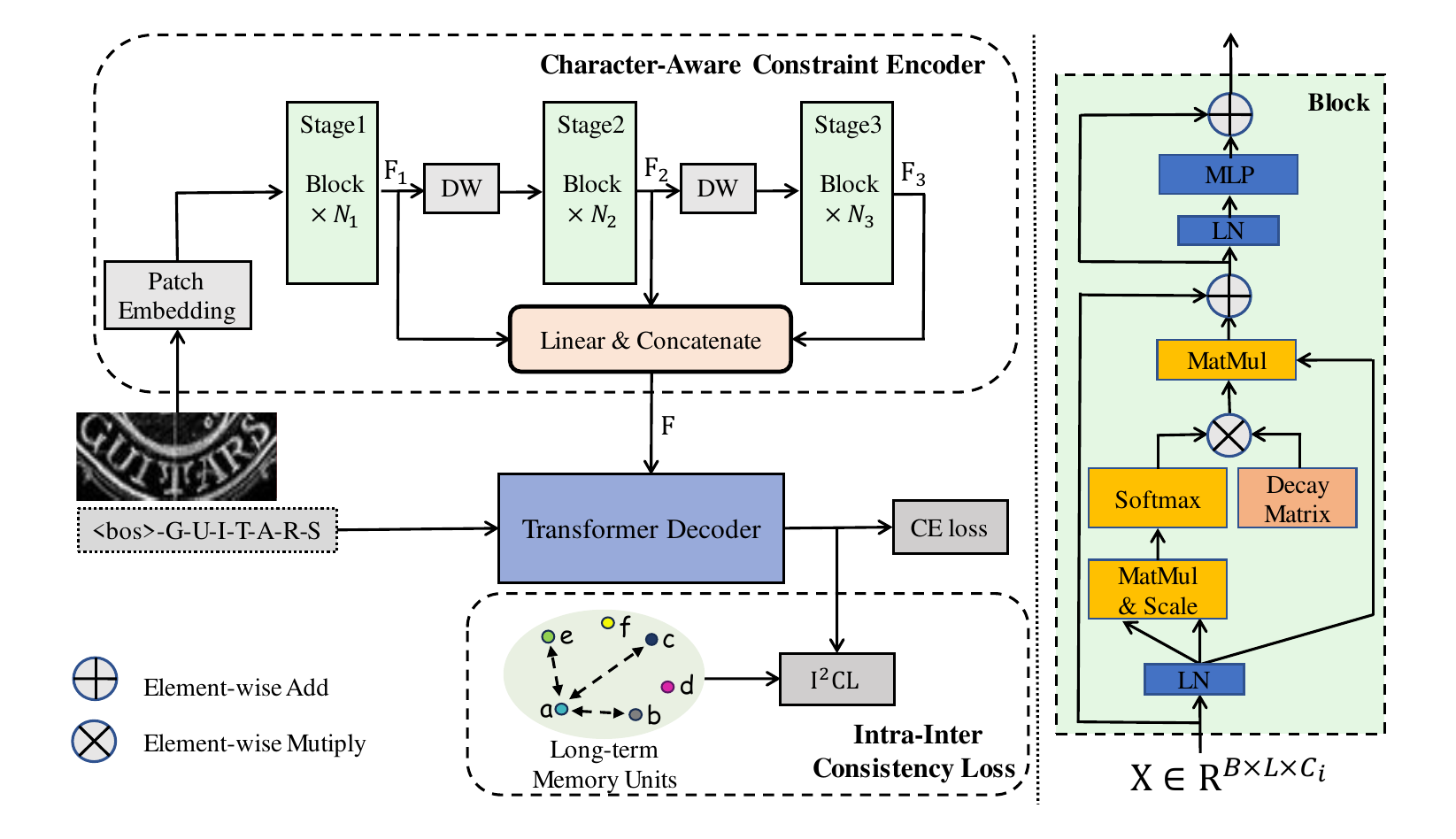}
\caption{
The framework of our CFE. The pipeline is composed of two key components: CACE and $\text{I}^{2}\text{CL}$. CACE explores the local patterns within character by utilizing the decay matrix. $\text{I}^{2}\text{CL}$ uses a set of learnable long-term memory units to represent the global character feature distribution in the decoding space.
CE loss denotes the cross entropy loss. DW means the 2x downsampling at height dimension using CNN.
}
\label{fig:pipeline}
\end{figure*}

\section{Related Work}
\subsection{Scene Text Recognition}
In scene text recognition models, the visual encoder is an essential component. 
It aims to provide discriminative visual feature representation for subsequent CTC~\cite{ctc} decoders, attention decoders, or Transformer decoders. 
Early methods employing CNN as encoder have been widely applied in various networks and applications~\cite{CRNN,aster,trba}. 
However, CNN-based approaches typically compress the height dimension of images into 1 during feature extraction, causing each visual feature to correspond to a thin-slice region in the image. 
This limitation leads to poor performance on irregular datasets. 
Recently, due to the significant advancements of Transformer in the visual domain, many recent works~\cite{vitstr,svtr,parseq,clipocr} have opted to use Vision Transformer as the visual encoder.
These methods demonstrate good performance on irregular datasets. 
ViTSTR~\cite{vitstr} utilizes the Vision Transformer as the encoder to model relationships between different visual tokens. 
SVTR~\cite{svtr} employs a pyramid-style Transformer as the visual encoder to guide the model in establishing global relationships between characters and local relationships within character. 
Although SVTR and ViTSTR are both pure visual models, there exist some performance differences. 
We attribute this to SVTR enhancing its ability to model the relationships between the character components.
For the problem of LISV, we also need to focus on the local features to obtain discriminative features for recognizing characters.
Therefore, we propose the Charatcer-Aware Constraint Encoder. 
It can perceive the local patterns within character by utilizing the decay matrix in each block. 

\subsection{Contrastive Learning in STR}
In scene text recognition, the optimized object is not only cross entropy or CTC loss~\cite{ctc}, but also combines contrastive learning to handle different problems. 
ConCLR~\cite{conclr} uses contrastive loss to bring identical characters closer and separate different characters in the embedding space. 
DiG~\cite{DiG} incorporates a contrastive learning branch to mimic human-like reading behavior and learn text discrimination. 
In the meantime, it employs a momentum branch to create a comprehensive and reliable dictionary on-the-fly. 
CornerTransformer~\cite{wordart} designs a Character Contrastive loss that implicitly learns common features for each character class for solving artistic text recognition. 
CLIP-OCR ~\cite{clipocr} introduces a linguistic consistency loss for aligning the intra-relationship and inter-relationship. 
We argue that contrastive learning is also helpful in solving the problems of LICV and SICV. 
Hence, we introduce the Intra-Inter Consistency Loss to take intra-class compactness and inter-class separability into account. 
This loss function can learn a long-term memory unit for each class, simultaneously ensuring that the memory units of different categories are far apart. 

\section{Proposed Method}
In this section, we first detail the pipeline of the proposed method CFE in Sec. \ref{sec:pipeline}, and then introduce Character-Aware Constraint Encoder (CACE) and Intra-Inter Consistency Loss ($\text{I}^{2}\text{CL}$) in Sec. \ref{sec:lemse} and Sec. \ref{sec:iicl} respectively.

\subsection{Pipeline}\label{sec:pipeline}
The pipeline of CFE is illustrated in Figure \ref{fig:pipeline} and it can be viewed as an encoder-decoder architecture.
Given a batch of images with size $B \times H \times W \times 3$, after three stages in CACE, we acquire the three visual sequences: $\mathbf{F_1} \in \mathbf{R}^{B \times \frac{HW}{16} \times C_1}$, $\mathbf{F_2} \in \mathbf{R}^{B \times \frac{HW}{32} \times C_2}$, $\mathbf{F_3} \in \mathbf{R}^{B \times \frac{HW}{64} \times C_3}$. 
Subsequently, we linearly map and concatenate them to get the final output visual sequences $\mathbf{F} \in \mathbf{R}^{B \times \frac{7}{64}HW \times C}$.
Next, $\mathbf{F}$ is fed into the Transformer Decoder to generate recognition features $\mathbf{O} \in \mathbf{R}^{B \times T \times C}$. Finally, we calculate the CE loss and $\text{I}^2\textbf{CL}$ separately and sum them up to obtain the training objective.

\subsection{Character-Aware Constraint Encoder}\label{sec:lemse}
To address the issue of LICV, we present a novel Character-Aware Constraint Encoder (CACE). 
In this encoder, the visual features are extracted from three stages, and each stage consists of multiple stacked blocks.
CACE introduces an explicit decay matrix $\mathbf{D}$ into the block to encode the local patterns ({\em e}.{\em g}., stroke, morphology, {\em etc}.) and relationships between the inter-character components. 
Additionally, to fully utilize the visual features output from the three stages, we employ a simple multi-scale fusion strategy to merge them. 

The block is depicted in Figure \ref{fig:pipeline}, we summarize how the block works in Eq \ref{eq:block}:
\begin{equation}\label{eq:block}
\begin{split}
    \mathbf{Q} &= (\text{LN}(\mathbf{X}))\mathbf{W_Q})\odot \Theta, \\ \mathbf{K} &=(\text{LN}(\mathbf{X}))\mathbf{W_K}) \odot \hat{\Theta}, \\ 
    \mathbf{V} &= \text{LN}(\mathbf{X}) \mathbf{W_V}, \\
    \mathbf{X} &= \mathbf{X} + (Softmax(QK^T/d) \odot \mathbf{D}) \mathbf{V},  \\
    \mathbf{X} &= \mathbf{X} + MLP(\text{LN}(X)),
\end{split}
\end{equation}
where $\mathbf{X} \in \mathbf{R}^{B \times L \times C_i}$ is the input of the block and LN stands for Layer Normalization. $\mathbf{W_Q}$, $\mathbf{W_k}$, $\mathbf{W_V }$ are the learnable projection matrices. 
$\Theta$ represents the position embedding and $\hat{\Theta}$ is its complex conjugate followed ~\cite{retentive}. 
$\mathbf{D}$ $\in \mathbf{R}^{B \times L \times L}$ represents the decay matrix which values in [0, 1]. 
We design three different ways to yield $\mathbf{D}$:
\begin{equation}\label{eq:decayoption}
\begin{split}
    D_{ij} = 
    \begin{cases}
    \gamma^{(|x_i - x_j| + |y_i - y_j|)},
    & \text{option 1} \\
    \gamma^{max(|x_i - x_j|, |y_i - y_j|)}, & \text{option 2} \\
    [|x_i - x_j| \leq w \ \& \ |y_i - y_j| \leq h], & \text{option 3}
    \end{cases}
\end{split}
\end{equation}
\begin{figure}[!t]
     \centering
 \includegraphics[width=0.9\linewidth]{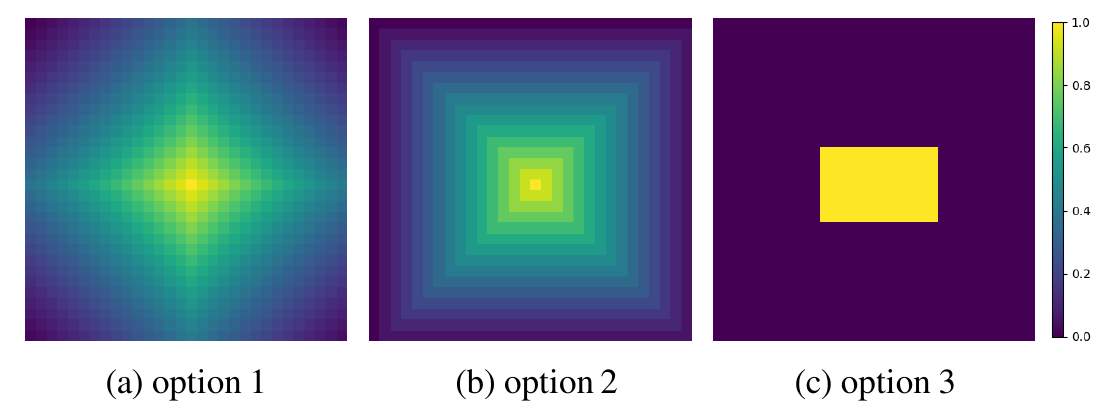}
     \caption{Visualization of three different options for generating decay matrix.
     }
     \label{fig:decayoption}
\end{figure}
where $x_i$ and $y_i$ denote the X and Y coordinates, respectively, of the $i$-th token in 2D space. $\gamma$ values range between 0 and 1~\cite{retentive}. 
[·] is an indicator function, if the condition is satisfied returning 1 else 0. 
$w$, $h$ are used to control the region during self-attention which are set 5 and 3 (insensitive) empirically. 
Further, we visualize $\mathbf{D}$ in the form of heatmap. 
As shown in Figure \ref{fig:decayoption}, the first two options correspond to dynamic decay, which is inversely proportional to the distances.
Option 3 employs fixed decay which utilizes a predefined binary window to control the attention region for each token like in SVTR ~\cite{svtr}. 
In this paper, we adopt option 2 as our choice. 
We argue that dynamic decay is compatible with human perception of distances.
Ablation study in Sec. \ref{sec:ablation} demonstrates its effectiveness.

In Eq. \ref{eq:decayoption}, $\mathbf{D}$ introduces spatial distances prior knowledge into the self-attention mechanism. 
The longer distances between token i and j, $\mathbf{D}_{i,j}$ become smaller.
Its purpose is to avoid the influence of irrelevant noisy tokens over long distances, allowing each token to model the different components within the whole character. 
Finally, the decay matrix enables the visual encoder to learn more distinctive visual representations for characters. 
Consequently, compared to global modeling without $\mathbf{D}$, it exhibits greater proficiency in capturing local patterns in each character. 
In practice, we employ $\mathbf{D}$ into the first x blocks for intra-character components modeling, while the remaining blocks do not use $\mathbf{D}$ for inter-character modeling.

To better capture the different patterns of a certain character, we design a fusion strategy to obtain multi-scale character features. 
Specifically, it fuses the features $\mathbf{F_1}$, $\mathbf{F_2}$, $\mathbf{F_3}$ by projecting them into the same hidden dimension. 
Overall, on the one hand, by adding the decay matrix, CACE enables each token to perceive local patterns within character, allowing for better extraction of discriminative features. 
On the other hand, through multi-scale fusion, CACE can integrate the local patterns of a character at different scales, which can enhance the diversity of character features.

\subsection{Intra-Inter Consistency Loss}\label{sec:iicl}
Although CACE can help alleviate the problem of LICV, the existence of SICV still prevents the model from reaching its optimal performance. 
The SICV issue leads to the phenomenon of mixture distribution between character features. 
Therefore, it is necessary to increase the distance between characters of different categories and decrease the distance between characters of the same category.
We believe that using the trait of contrastive learning to solve the SICV problem is a good choice.
However, existing contrastive learning methods mostly cluster characters within a batch, lacking the perception of global character feature distribution. 
Hence, we introduce the Intra-Inter Consistency Loss ($\text{I}^{2}\text{CL}$) to explore the distribution of each character category by learning a set of long-term memory units.

Formally, the $\text{I}^{2}\text{CL}$ is defined as illustrated in Eq \ref{eq:ccloss}:
\begin{equation}\label{eq:ccloss}
\begin{split}
    L_{cl} = \frac{1}{2} \sum_{i=1}^{BT}\frac{||O_i - c_{y_i}||_2^{2}}{(\Sigma_{j=1, j \neq y_i}^{V}||O_i - c_j||_2^{2}) + \delta},
\end{split}
\end{equation}
where $L_{cl}$ represents the $\text{I}^2\text{CL}$. Initially, we reshape the output feature $\text{O}$ into $\mathbf{R}^{BT \times C}$. 
$BT$ denotes the number of characters in a batch and $T$ is the maximum length of character sequences in a text. 
$\text{O}_i \in \mathbf{R}^{C}$ signifies the $i$-th character feature, $y_i$ denotes the label of $\text{O}_i$. 
$c_{y_i}$ represents the memory unit of the $y_i$-th character category in decoding space which can be updated during training. 
$V$ denotes the vocabulary size. 
$\delta$ is a constant used to prevent the denominator from equaling 0 and we set $\delta = 1$ by default.

Compared with DiG~\cite{DiG} and CornerTransformer~\cite{wordart}, although these approaches all use contrastive learning loss, there remain some differences.
DiG and CornerTransformer only consider the intra-class compactness and inter-class separability of features in a batch. 
But $\text{I}^{2}\text{CL}$ considers the clustering between all the training samples. 
After training, the memory units of our $\text{I}^2\text{CL}$ will be a discrete distribution because there is a penalty for too small distance between different memory units.
In addition, we can use these memory units to represent the global character feature distribution.
From one perspective, this discrete distribution will increase the distance between characters of different categories and solve the SICV problem. 
From another perspective, all characters will be tightly distributed around the memory units corresponding to their categories, alleviating the LICV problem. 

\subsection{Training Objective}
The final objective function of the proposed method is formulated in Eq. \ref{eq:loss}. $L_{ce}$ represents the cross entropy loss. 
$\lambda$ denotes the scalar used to balance the two loss functions and is set to 0.2.
\begin{equation}\label{eq:loss}
   L = L_{ce} + \lambda L_{cl}. 
\end{equation}

\begin{table*}[!t]
\centering
\renewcommand{\arraystretch}{0.6}
\resizebox{0.8\linewidth}{!}{
\begin{tabular}{c|c|c|c|c|c|c}
\toprule
Models & [$C_1,C_2,C_3$] & [$N_1,N_2,N_3$] & Heads & $C$ & Decay Order & Params(M) \\ \midrule
CFE-Tiny & [64, 128, 256] & [3,6,3] & [2,4,8] & 128 & 6-6 & 4.5 \\
CFE-Small & [96, 192, 256] & [3,6,6] & [3,6,8] & 192 &
8-7& 9.2\\
CFE-Base & [128, 256, 384] & [3,6,9] & [4,8,12] & 256 &
8-10  & 23.9 \\ \bottomrule
\end{tabular}}
\caption{Various configurations of our CFE. Heads denote the number of attention heads used in each stage. For Decay Order x-y, x signifies using $\mathbf{D}$ in the first x blocks, while y denotes not using $\mathbf{D}$ in the last y blocks. Params(M) indicates the trainable parameters of the model.}
\label{tab:params}
\end{table*}

\begin{table*}[!t]
\centering
\renewcommand{\arraystretch}{1.22}
\resizebox{\linewidth}{!}{
\begin{tabular}{@{}l|cccccc|c|ccccccc|c|c}
\toprule
\multirow{2}{*}{Method} & \multicolumn{7}{c|}{Common Benchmarks} & \multicolumn{8}{c|}{Union14M-Benchmark} & P \\ \cline{2-16}
 & IC13 & SVT & IIIT & IC15 & SVTP & CT &
WAVG & Cur & M-O & Art & Con & Sal & M-W & Gen & AVG & (M) \\ \midrule
$\text{RobustScanner}^{*}$~\cite{robustscanner} & 94.8     & 88.1    & 95.3      & 77.1     & 79.5     & 90.3       & 88.4  & 43.6 & 7.9 & 41.2 & 42.6 & 44.9 & 46.9 & 39.5 & 38.1 & -  \\ 
$\text{PARSeq}_A^{\#}$~\cite{parseq} & 97.0 & 93.6 & 97.0 & 86.5 & 88.9 & 92.3 & 93.3 &58.2 & 17.2 & 54.2 & 59.4 & 67.7 & 55.8 & 61.1 & 53.4 & 23.8 \\ 
$\text{SVTR}^{*}$~\cite{svtr}  & 97.1 & 91.5 & 96.0 & 85.2 & 89.9 & 91.7 & 92.3 & 63.0 & \textbf{32.1} & 37.9 & 44.2 & 67.5 & 49.1 & 52.8 & 49.5 &  24.6 \\ 
$\text{CLIP-OCR}^{\dag}$~\cite{clipocr} & 97.7 & 94.7 & 97.3 & 87.2 & 89.9 & 93.1 & 93.8 & 59.4  & 15.9 & 57.6 & 59.2 & 69.2 & 62.6 & 62.3 & 55.2 & 31.1 \\
$\text{VisionLAN}^{*}$~\cite{visionlan} & 95.7 & 91.7 & 95.8 & 83.7 & 86.0 & 88.5 & 91.2 & 57.7 & 14.2 & 47.8 & 48.0 & 64.0 & 47.9 & 52.1 & 47.4 & 32.8 \\
$\text{ABINet}^{*}$~\cite{abinet} & 97.4  & 93.5  & 96.2 & 86.0 & 89.3 & 89.2  & 92.3 & 59.5 & 12.7 & 43.3 & 38.3 & 62.0 & 50.8 & 55.6 & 46.0 & 36.7 \\
$\text{MATRN}^{*}$~\cite{matrn} & \textbf{97.9} & \textbf{95.0} & 96.6 & 86.6 & 90.6 & 93.5 & 93.5  & 63.1 & 13.4 & 43.8 & 41.9 & 66.4 & 53.2 & 57.0 & 48.4 & 44.2\\ 
$\text{SRN}^{*}$~\cite{SRN} & 95.5  & 91.5  & 94.8  & 82.7   & 85.1    & 87.8  & 90.4  & 63.4 & 25.3 & 34.1 & 28.7 & 56.5 & 26.7 & 46.3 & 39.6 & 54.7 \\
$\text{MGP-Base}^{\dag}$~\cite{MGP} & 97.3 & 94.7 & 96.4 & 87.2 & 91.0 & 90.3 & 93.3 & 55.2 & 14.0 & 52.8 & 48.4 & 65.1 & 48.1 & 59.0 & 48.9 & 148.0 \\
 \midrule
$\text{LPV-Tiny}^{\dag}$~\cite{zhangbo} & 96.7 & 92.9 & 96.3 & 86.4 & 86.7 & 90.6 & 92.5 & 55.0 & 11.7 & 51.4 & 53.9 & 65.6 & 52.1 & 58.1 & 49.7 & 8.1 \\
$\text{LPV-Small}^{\dag}$~\cite{zhangbo} & 96.8 & 93.7 & 96.7 & 87.1 & 89.8 & 92.4 & 93.3 & 61.5 & 15.7 & 55.9 & 58.8 & 69.0 & 62.1 & 60.3 & 54.8 & 14.0 \\
$\text{LPV-Base}^{\dag}$~\cite{zhangbo} & 97.6 & 94.6 & 97.3 & \textbf{87.5} & 90.9 & 94.8 & 94.0 & 68.3 & 21.0 & 59.6 & 65.1 & \textbf{76.2} & 63.6 & 62.0 & 59.4 & 35.1 \\ \midrule
$\text{LISTER-Tiny}^{\dag}$~\cite{lister} & 97.7 & 93.5 & 96.5 & 86.5 & 87.8 & 87.9 & 92.8 & 49.3 & 13.2 & 49.1 & 58.2 & 58.2 & 64.2 & 60.8 & 50.4 & 19.9 \\
$\text{LISTER-Base}^{\dag}$~\cite{lister} & \textbf{97.9} & 93.8 & 96.9 & \textbf{87.5} & 89.6 & 90.6 & 93.5 & 54.8 & 17.2 & 51.3 & 61.5 & 62.6 & 61.3 & 62.9 & 53.1 & 49.9\\ \midrule
CFE-Tiny(Ours) & 96.7 & 93.5 & 96.9 & 85.7 & 86.8 & 91.7 & 92.7 & 56.6 & 14.0 & 53.6 & 64.7 & 67.0 & 63.6 & 61.4 & 54.4 & 4.5\\
CFE-Small(Ours) &  96.7 & 93.7 & 97.2 & 86.8 & 89.9 & 93.1 & 93.4 & 63.4 & 17.3 & 57.5 & 71.5 & 73.2 & 64.2 & 63.8 & 58.7 & 9.2 \\
CFE-Base(Ours) & 97.6 & 94.3 & \textbf{97.9} & 86.9 & \textbf{91.8} & \textbf{95.5} & \textbf{94.1} & \textbf{70.0} & 20.8 & \textbf{62.4} & \textbf{72.0} & 75.2 & \textbf{65.7} & \textbf{65.1} & \textbf{61.6} & 23.9 \\
\bottomrule
\end{tabular}}
\caption{Performance of models trained on synthetic datasets. * means the results on Union14M-Benchmark is from MAERec~\protect\cite{union14m}. $\dag$ signifies we use the released checkpoints to test Union14M-Benchmark. $\#$ implies we retrain the model on the synthetic datasets and then test the result on Union14M-Benchmark. Cur, M-O, Art, Ctl, Sal, M-W, and Gen respectively represent Curve, Multi-Oriented, Artistic, Contextless, Salient, Multi-Words, and General.  For simplicity, they have the same meaning in the following. P(M) indicates the trainable parameters. Bold values denote the first accuracy in each column.}
\label{tab:performance}
\end{table*}

\section{Experiment}
\subsection{Datasets}
Following the setup of \cite{MGP,abinet}, we conduct experiments using MJSynth~\cite{MJ1,MJ2} and SynthText~\cite{ST} as training data. 
The training data consists of 16M synthetic text images. 
We evaluate the model on 6 common benchmarks containing IIIT ~\cite{IIIT}, IC13~\cite{IC13}, IC15~\cite{IC15}, SVT~\cite{SVT}, SVTP~\cite{SVTP} and CT~\cite{CUTE}. 
To further valid the effectiveness of our CFE, we extra test performance on Union14M-Benchmark~\cite{union14m}, ArT~\cite{art}, COCO-Text~\cite{cocotext}, Uber-Text~\cite{uber}, and WordArt~\cite{wordart}. 
Beyond that, we also supply a few experiments trained on Union14M-L.

\subsection{Implementation Details}
To balance the accuracy and speed, we develop three variations with varying numbers of parameters similar to SVTR and LPV~\cite{zhangbo}. 
The specific network configurations are detailed in Table \ref{tab:params}.
The images are resized to $32 \times 128$. For data augmentation, we follow the random data augmentation methods~\cite{parseq} including Invert, GaussianBlur, Sharpness, and PoissonNoise. 
The vocabulary size $V$ is 96, which comprises mixed-case alphanumeric characters, punctuation marks, [BOS] for the beginning symbol, and [EOS] for the ending symbol. 
The maximum label length $T$ is set to 25. 
We train 20 epochs with a warm-up of 1.5 epochs, utilizing the Adam optimizer~\cite{adam} with a learning rate of 5e-4. 
We add $L_{cl}$ in the last 25\% time of the training process. 
By this point, the model has already demonstrated convincing recognition capabilities, allowing more accurate learning for memory units and clustering. 
We use the Transformer Decoder with 1 layer as the recognition decoder and adopt autoregressive decoding for training.
The experiments are conducted on 4 NVIDIA 4090 GPUs with a batch size of 384.

\subsection{Evaluation Metric}
For validation, we configure the vocabulary size to 36, encompassing 0-9 and a-z. 
We employ the word accuracy as the evaluation metric. 
Consistent with~\cite{fewlabels}, we record the weighted average score (WAVG) on common benchmarks based on the number of samples.
As for Union14M-Benchmark, we report the average score (AVG) like~\cite{union14m}.

\subsection{Comparisons with State-of-the-Arts}
In Table \ref{tab:performance}, we compare our CFE with multiple recent state-of-the-art methods on common benchmarks. 
All the methods are trained by synthetic image texts for fair comparison. 
CFE-Base shows significant performance, especially in datasets of regular IIIT and irregular SVTP and CT.   It achieves SOTA accuracy of 94.1\% while keeping low parameters (23.9M) compared to other STR models. 
Further, CFE-Tiny outperforms LPV-Tiny with only 4.5M parameters, and CFE-Small acquires a good performance of 93.4\% with 9.2M parameters. 
Besides, CFE-Base achieves SOTA performance on Union14M-Benchmark, with an average score of 61.6\%. 
For challenging datasets in Table \ref{tab:challenge}, although CFE-Base attains first or second accuracy, there remains a large gap with LISTER-Base on the Uber-Text dataset which contains many vertical texts. 
This is because LISTER-Base performs a rotation operation based on the original aspect ratio of the images, enhancing its ability to recognize vertical texts. 

The performance trained on Union14M-L is shown in Table \ref{tab:unionperformance}. 
Compared to the recent methods with additional linguistic information, our CFE-Base can continue to keep the best accuracy on most datasets of Union14M-Benchmark. 
These results all imply that our CFE is effective in enhancing the discriminative character features for solving challenging scene text recognition and can maintain superior performance in recognizing simple texts.

\begin{table}[!t]
\centering
\renewcommand{\arraystretch}{1.22}
\resizebox{0.85\linewidth}{!}{
\begin{tabular}{@{}c|c|c|c|c}
\toprule
Method & ArT & COCO & Uber & WordArt \\ \midrule
$\text{ABINet}^{\dag}$ ~\cite{abinet} & 65.4 & 57.1 & 34.9 & 67.4 \\
$\text{PARSeq}_A$ ~\cite{parseq} & 70.7 & 64.4 &42.0 & - \\
ConnerTransfomer ~\cite{wordart} & - & - & - & 70.8 \\
$\text{MGP-Base}^{\dag}$ ~\cite{MGP} & 69.0 & 65.4 & 40.7 & 72.4 \\
$\text{LISTER-Tiny}^{\dag}$ ~\cite{lister} & 69.0 & 64.1 & 48.0 & 67.6 \\
$\text{LISTER-Base}^{\dag}$ ~\cite{lister} & 70.1 & 65.8 & \textbf{49.0} & 69.8 \\
$\text{CLIP-OCR}$ ~\cite{clipocr} & 70.5 & \textbf{66.5} & 42.4 & 73.9 \\ 
\midrule
CFE-Tiny(Ours) & 69.8 & 62.8 & 42.3 & 71.0 \\
CFE-Small(Ours) & 71.8 & 64.2 & 42.8 & 74.0 \\
CFE-Base (Ours) & \textbf{72.8} & 66.3 & 43.6 & \textbf{75.7} \\
\bottomrule
\end{tabular}}
\caption{Comparison with SOTA methods on challenging
datasets. $\dag$ means to test performance using open-released model weights.
}

\label{tab:challenge}
\end{table}
\begin{table}[!t]
\centering
\renewcommand{\arraystretch}{0.9}
\resizebox{0.8\linewidth}{!}{
\begin{tabular}{@{}l|ccccccc|c}
\toprule
\multirow{2}{*}{Method} & Cur & M-O & Art & Con \\
 & Sal & M-W & Gen & AVG  \\ \midrule
\multirow{2}{*}{$\text{MGP-Base}^{\#}$ ~\cite{MGP}} & 78.8 & 74.3 & 67.7 & 68.7 \\
 & 75.7 & 60.0 & 80.1 & 72.2 \\ \midrule
\multirow{2}{*}{$\text{LPV-Base}^{\#}$ ~\cite{zhangbo}} & 85.2 & 75.9 & 74.8 & 80.5 \\
 & 83.3 & 82.2 & 82.8 & 80.7 \\ \midrule
\multirow{2}{*}{$\text{CLIP-OCR}^{\#}$ ~\cite{clipocr}} & 84.6 & \textbf{83.1} & 76.3 & 80.0 \\
& 81.3 & 81.8 & 83.9 & 81.6 \\ \midrule
\multirow{2}{*}{$\text{LISTER-Base}^{\#}$ ~\cite{lister}} & 70.9 & 51.1 & 65.4 & 73.3 \\
& 66.9 & 77.6 & 77.9 & 69.0 \\ \midrule
\multirow{2}{*}{MAERec-Base ~\cite{union14m}} & 76.5 & 67.5 & 65.7 & 75.5 \\
 & 74.6 & 77.7 & 81.8 & 74.2 \\ \midrule
\multirow{2}{*}{CFE-Tiny} & 77.3 & 62.1 & 69.6 & 79.1 \\
& 74.2 & 77.9 & 79.9 & 
 74.3 \\ \midrule
\multirow{2}{*}{CFE-Small} & 84.4 & 73.4 & 75.4 & 84.7 \\
& 81.3 & 83.3 & 82.8 & 80.8 \\ \midrule
\multirow{2}{*}{CFE-Base} & \textbf{86.8} & 80.4  & \textbf{77.5}  & \textbf{85.5} \\
& \textbf{83.5} & \textbf{85.9} & \textbf{84.4} & \textbf{83.4} \\ 
\bottomrule
\end{tabular}
}
\caption{Performance on models trained on Union14M-L. $\#$ implies we retrain the model on Union14M-L and then test on Union14M-Benchmark.}
\label{tab:unionperformance}
\end{table}

\begin{table}[!t]
\centering
\renewcommand{\arraystretch}{1.22}
\resizebox{\linewidth}{!}{
\begin{tabular}{c|c|c|c|c|c|c|c|c|c}
\toprule
CACE & $\text{I}^2\text{CL}$ & Cur & M-O & Art & Ctl & Sal      & M-W  & Gen &   AVG   \\  \midrule
-  &  -  & 68.5 & 20.2 & 58.6 &  66.9 & 74.4 & 65.4 & 64.7  &  59.8   \\ 
\checkmark & -  & 68.3 & 21.4 & 59.8 & 70.1 & 74.8 & 66.1 & 64.9 & 60.8  \\ 
-  & \checkmark & 67.6 & 20.7 & 59.4 & 71.4 & 76.0 & 64.2 & 65.2 & 60.6 \\ 
\checkmark    & \checkmark   & 
70.0 & 20.8 & 62.4 & 72.0  & 75.2 & 65.7 & 65.1 & \textbf{61.6} \\ 
\bottomrule
\end{tabular}
}
\caption{The effectiveness of CACE and $\text{I}^2\text{CL}$.}
\label{tab:lemseiicl}
\end{table}

\begin{table}[!t]
\centering
\renewcommand{\arraystretch}{1.22}
\resizebox{\linewidth}{!}{
\begin{tabular}{c|c|c|c|c|c|c|c|c}
\toprule
Method & Cur & M-O & Art & Ctl & Sal  & M-W  & Gen & AVG \\ \midrule
- & 68.3 & 21.4 & 59.8 & 70.1 & 74.8 & 66.1 & 64.9 & 60.8 \\ 
CC loss & 66.4 & 19.1 & 58.0 & 69.2 & 73.6 & 68.0 & 64.8 & 59.9 \\ 
$\text{I}^2\text{CL}$  & 70.0 & 20.8 & 62.4 & 72.0 & 75.2 & 65.7 & 65.1 & \textbf{61.6} \\ \bottomrule
\end{tabular}}
\caption{Comparision with other contrastive loss.}
\label{tab:contrastiveloss}
\end{table}

\begin{table}[!t]
\centering
\renewcommand{\arraystretch}{1.22}
\resizebox{\linewidth}{!}{
\begin{tabular}{c|c|c|c|c|c|c|c|c|c}
\toprule
$\mathbf{D}$ & M-S & Cur & M-O & Art & Ctl & Sal  & M-W  & Gen &   AVG   \\  \midrule
-  &  -  &  67.6 & 20.7 & 59.4 & 71.4 & 76.0 & 64.2 & 65.2 & 60.6 \\ 
\checkmark & - & 68.0 & 20.7 & 58.5 & 73.3 & 75.5 & 66.6 & 65.2 & 61.1 \\ 
-  & \checkmark & 66.3 & 18.9 & 58.8 & 70.7 & 73.2 & 68.7 & 64.7 & 60.2 \\ 
\checkmark  & \checkmark  & 70.0 & 20.8 & 62.4 & 72.0 & 75.2 & 65.7 & 65.1 & \textbf{61.6} \\ \bottomrule
\end{tabular}}
\caption{The effectiveness of decay matrix $\mathbf{D}$ and multi-scale fusion in CACE. M-S represents the multi-scale fusion.}
\label{tab:decaymatrix-multisacle}
\end{table}

\subsection{Ablation Study}\label{sec:ablation}
\subsubsection{The Effectiveness of CACE and $\text{I}^2\text{CL}$}
To investigate the effectiveness of CACE and $\text{I}^2\text{CL}$, we conduct experiments in Table \ref{tab:lemseiicl}. 
The baseline refers to not using CACE and $\text{I}^2\text{CL}$ which achieves an accuracy of 59.8\%.  
After adding CACE to guide the encoder to perceive the morphological information and character region, we can get an average accuracy of 60.8\% (+1.0\%). 
This improvement implies that the encoder can model the local patterns of each character by using $\mathbf{D}$ and multi-scale fusion. 
On the other hand, only adopting $\text{I}^2\text{CL}$ obtains the accuracy of 60.6\% (+0.8\%) which means learning long-term memory units is effective for solving LISV and SICV. 
When employing them together, we further improve the performance to 61.6\% (+1.8\%). 
These results demonstrate that our CFE can utilize the enriched features to enhance the discriminability of characters and that CACE and $\text{I}^2\text{CL}$ can cooperate well to optimize our model. 

Moreover, Table \ref{tab:contrastiveloss} compares our $\text{I}^2\text{CL}$ with the Character Contrastive loss (CC loss) designed by CornerTransformer. 
The result indicates that our $\text{I}^2\text{CL}$ surpasses the conventional contrastive learning method CC loss. 
Compared to the performance without contrastive learning loss, the introduction of CC loss leads to a decrease of 0.9\% (59.9\%).
We argue that achieving small intra-class differences and large inter-class differences is impractical by clustering characters only within a batch.
Hence, CC loss fails to learn the global feature distribution of each category.
In contrast, our $\text{I}^2\text{CL}$ employs the trainable long-term memory units to represent the global distribution of each category, aiding the model in better achieving intra-class compactness and inter-class separability. 
This verifies that our $\text{I}^2\text{CL}$ can enhance the global character feature distribution for solving challenging STR.

\subsubsection{The Components of CACE}
To evaluate CACE, we compare the results with decay matrix $\mathbf{D}$ or multi-scale fusion in 
Table \ref{tab:decaymatrix-multisacle}. 
The first row which only relies on $L_{cl}$ for training, obtains a performance of 60.6\%. 
When introducing $\mathbf{D}$ in the block, the model exhibits 0.5\% (61.1\%) improvement. 
The inclusion of $\mathbf{D}$ enables the encoder to perceive the local patterns within the character and generate discriminative features for recognizing characters. 
However, when only adding multi-scale fusion, CFE-Base experiences a decrease of 0.4\% (60.2\%). 
We argue that relying solely on multi-scale fusion leads the model to ignore the local character-level features, decreasing performance on challenging datasets. 
A total improvement of 1.0\% (61.6\%) is achieved when employing them together. 
We attribute the improvement to two reasons: 1) Decay matrix allows the encoder to perceive the morphological information at character level, helping to distinguish the characters with different appearances. 
2) Multi-scale fusion provides diverse character features by integrating the features from the three stages.

\begin{table}[!t]
\centering
\renewcommand{\arraystretch}{1.22}
\resizebox{\linewidth}{!}{
\begin{tabular}{c|c|c|c|c|c|c|c|c}
\toprule
Option & Cur & M-O & Art & Ctl & Sal  & M-W  & Gen &   AVG    \\ \midrule
1  &  67.1 & 19.6 & 59.8 & 71.6 & 75.4 & 71.2 & 65.0 & 61.4 \\ 
2  & 70.0 & 20.8 & 62.4 & 72.0 & 75.2 & 65.7 & 65.1 & \textbf{61.6} \\ 
3 & 68.3 & 19.8 & 56.8 & 69.2 & 74.3 & 67.2 & 64.4 & 60.0 \\ \bottomrule
\end{tabular}
}
\caption{Different options of decay matrix.}
\label{tab:gendecay}
\end{table}

\begin{figure}[!t]
     \centering
     \includegraphics[width=0.85\linewidth]{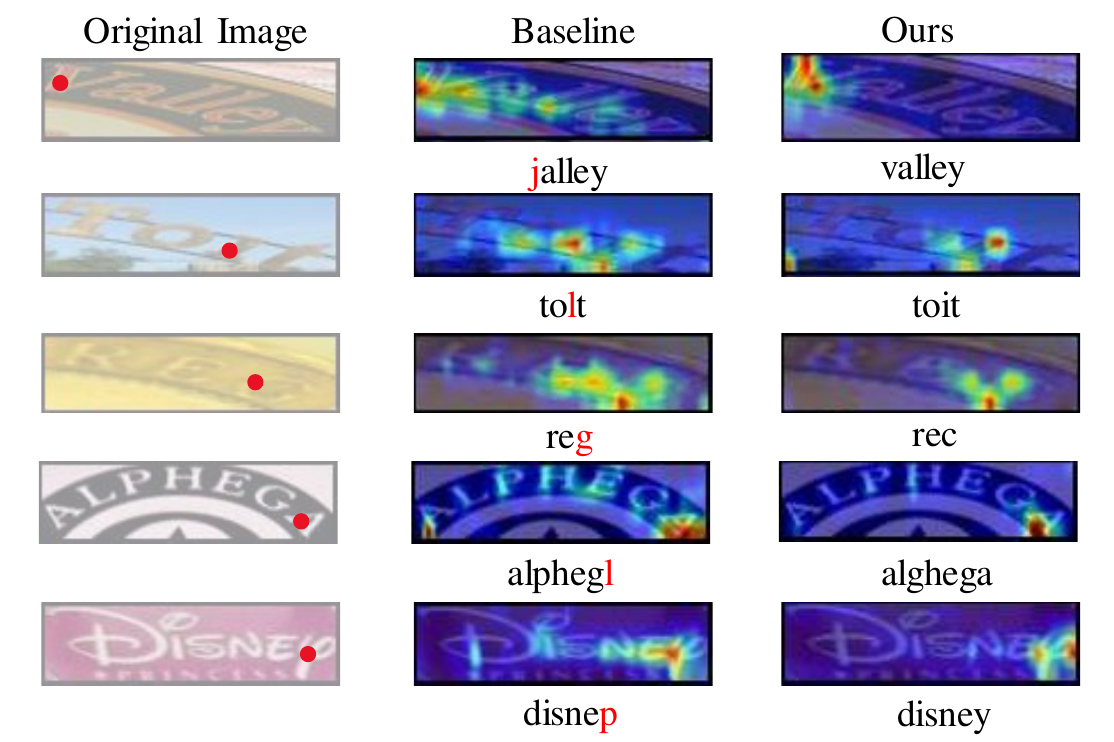}
     \caption{Visualization of attention maps in CACE. In the first column, the red point in each image is the query token. The second column images imply we use the baseline to calculate the attention scores between the red point and all points. The third column images mean CFE is used to calculate the attention scores.
     }
    \label{fig:attenmaps}
\end{figure}

\begin{figure}[!t]
     \centering
\includegraphics[width=1\linewidth]{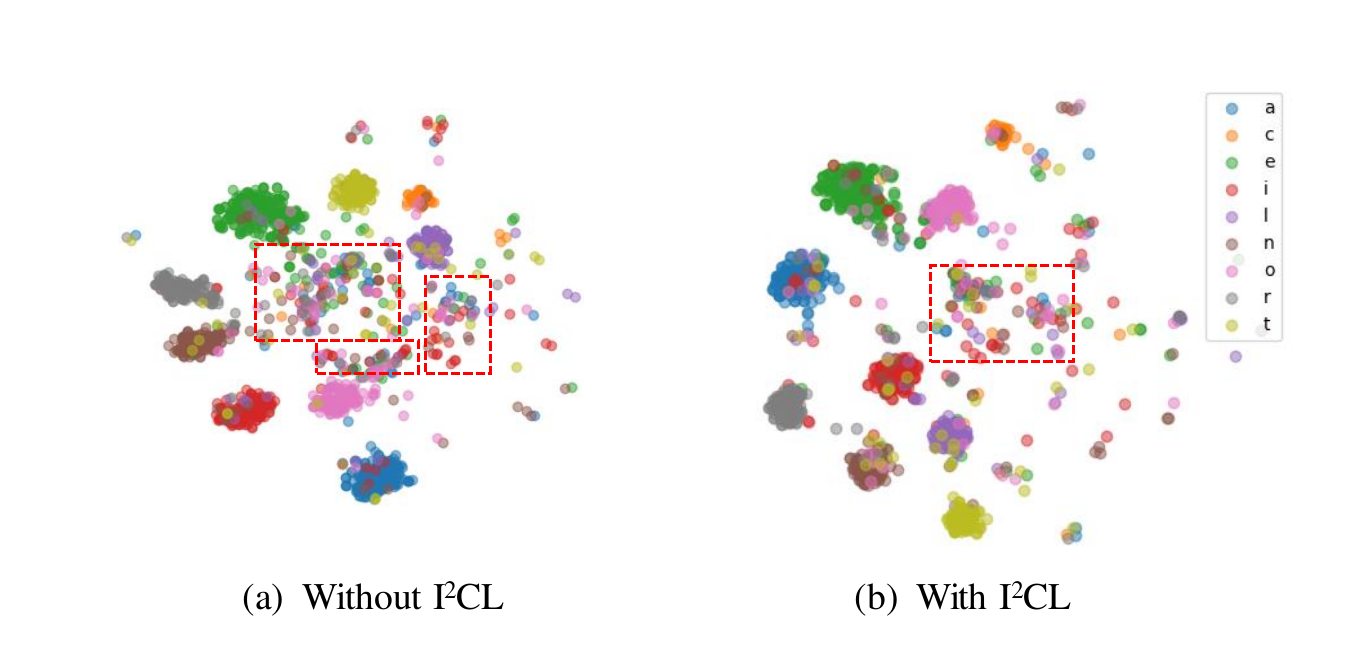}
     \caption{Visualization of character feature distribution. The feature points in the red rectangle mean the mixture distribution. Zoom in for better visualization.
     }
     \label{fig:visualization}
\end{figure}

\subsubsection{The Different Options of Decay Matrix $\mathbf{D}$}
In Table \ref{tab:gendecay}, we study different options of yielding $\mathbf{D}$ and find that option 2 achieves the best performance. 
When considering the tokens i and j are components of a character, option 2 generates a larger $\textbf{D}_{i,j}$. So option 2 can focus more on the character region than option 1. 
The accuracy rates of option 2 is 61.6\%, slightly higher than that of option 1 (61.4\%).
However, option 3 performs less competitively than options 1 and 2 with 60.0\% accuracy.
The reason is that dynamic decay will generate more precise attention region for each token.

\subsection{Visualization and Analysis}\label{sec:visualization}
\subsubsection{The Visualization of CACE}
From the perspective of understanding how to focus on the character region, we visualize the self-attention scores in Figure \ref{fig:attenmaps}. 
Specifically, we use the token that corresponds to the red point as the query to calculate the attention scores with all visual tokens and then reshape them to 2D.
For the first three rows, while focusing on the character region, the baseline also interacts with nearby characters, leading to recognition errors.
This problem can be put down to insufficient exploration of the local patterns. 
In contrast, CACE recognizes the texts correctly by concentrating exclusively on the character region. 
For the last two rows, the baseline is prone to interference from distant noise, ultimately causing errors. 
Conversely, CACE avoids interference from distant noise, ensuring accurate recognition.
Through this, we conclude that CACE can perceive the region of each character and then model the components of the character.

\subsubsection{The Qualitative Analysis of $\text{I}^2\text{CL}$}
To evaluate the effectiveness of our $\text{I}^2\text{CL}$ and justify its design, we use t-SNE~\cite{tsne} to reduce the feature dimension to 2D space for visualization.
Figure \ref{fig:visualization} illustrates the feature distribution of 9 easily misrecognized characters.
In Figure \ref{fig:visualization}(a) without $\text{I}^2\text{CL}$, we observe that the distribution of different category characters is severely mixed, as symbolized in the red rectangle. After introducing $\text{I}^2\text{CL}$, the phenomenon of mixed distribution can be alleviated. 
In addition, the distance of characters between different categories is widened as shown in Figure \ref{fig:visualization}(b). 
By training a discrete distribution for long-term memory units, all characters of different categories can be compactly distributed around the memory units, while ensuring the separability between classes. 
These phenomena prove the effectiveness of $\text{I}^2\text{CL}$ and are consistent with our design.

\section{Conclusion}
In this paper, we notice the two problems of Large Intra-Class Variance (LICV) and Small Inter-Class Variance (SICV) in challenging texts. To address these issues, a Character Features Enriched model (CFE) is proposed to obtain the discriminative features via Character-Aware Constraint Encoder (CACE) and Intra-Inter Consistency Loss ($\text{I}^2\text{CL}$). 
Firstly, CACE enables visual tokens to perceive character morphological information by introducing the decay matrix, which enhances the discriminability of character features.
Secondly, $\text{I}^2\text{CL}$ helps achieve intra-class compactness and inter-class separability by learning a discrete distribution for long-term memory units.
The experimental results show that CFE can not only effectively improve the performance on challenging texts, but also maintain high accuracy on simple texts, taking a further step toward STR with strong robustness.

\section*{Acknowledgments}
This work is supported by the National Key Research and Development Program of China (2022YFB3104700), the National Nature Science Foundation of China (62121002, U23B2028, 62102384).
This research was support by the Supercomputiong Center of the USTC. 
We acknowledge the support of GPU cluster built by MCC Lab of Information Science and Technology Institution, USTC.

\bibliographystyle{named}
\bibliography{ijcai24}

\begin{thebibliography}{}

\bibitem[\protect\citeauthoryear{Atienza}{2021}]{vitstr}
Rowel Atienza.
\newblock Vision transformer for fast and efficient scene text recognition.
\newblock In {\em International Conference on Document Analysis and Recognition}, pages 319--334. Springer, 2021.

\bibitem[\protect\citeauthoryear{Baek \bgroup \em et al.\egroup }{2019}]{trba}
Jeonghun Baek, Geewook Kim, Junyeop Lee, Sungrae Park, Dongyoon Han, Sangdoo Yun, Seong~Joon Oh, and Hwalsuk Lee.
\newblock What is wrong with scene text recognition model comparisons? dataset and model analysis.
\newblock In {\em International Conference on Computer Vision (ICCV)}, 2019.

\bibitem[\protect\citeauthoryear{Baek \bgroup \em et al.\egroup }{2021}]{fewlabels}
Jeonghun Baek, Yusuke Matsui, and Kiyoharu Aizawa.
\newblock What if we only use real datasets for scene text recognition? toward scene text recognition with fewer labels.
\newblock In {\em Proceedings of the IEEE/CVF Conference on Computer Vision and Pattern Recognition}, pages 3113--3122, 2021.

\bibitem[\protect\citeauthoryear{Bautista and Atienza}{2022}]{parseq}
Darwin Bautista and Rowel Atienza.
\newblock Scene text recognition with permuted autoregressive sequence models.
\newblock In {\em European Conference on Computer Vision}, pages 178--196. Springer, 2022.

\bibitem[\protect\citeauthoryear{Cheng \bgroup \em et al.\egroup }{2023}]{lister}
Changxu Cheng, Peng Wang, Cheng Da, Qi~Zheng, and Cong Yao.
\newblock Lister: Neighbor decoding for length-insensitive scene text recognition.
\newblock In {\em Proceedings of the IEEE/CVF International Conference on Computer Vision}, pages 19541--19551, 2023.

\bibitem[\protect\citeauthoryear{Chng \bgroup \em et al.\egroup }{2019}]{art}
Chee~Kheng Chng, Yuliang Liu, Yipeng Sun, Chun~Chet Ng, Canjie Luo, Zihan Ni, ChuanMing Fang, Shuaitao Zhang, Junyu Han, Errui Ding, et~al.
\newblock Icdar2019 robust reading challenge on arbitrary-shaped text-rrc-art.
\newblock In {\em 2019 International Conference on Document Analysis and Recognition (ICDAR)}, pages 1571--1576. IEEE, 2019.

\bibitem[\protect\citeauthoryear{Du \bgroup \em et al.\egroup }{2022}]{svtr}
Yongkun Du, Zhineng Chen, Caiyan Jia, Xiaoting Yin, Tianlun Zheng, Chenxia Li, Yuning Du, and Yu-Gang Jiang.
\newblock Svtr: Scene text recognition with a single visual model.
\newblock {\em arXiv preprint arXiv:2205.00159}, 2022.

\bibitem[\protect\citeauthoryear{Fan \bgroup \em et al.\egroup }{2023}]{rmt}
Qihang Fan, Huaibo Huang, Mingrui Chen, Hongmin Liu, and Ran He.
\newblock Rmt: Retentive networks meet vision transformers.
\newblock {\em arXiv preprint arXiv:2309.11523}, 2023.

\bibitem[\protect\citeauthoryear{Fang \bgroup \em et al.\egroup }{2021}]{abinet}
Shancheng Fang, Hongtao Xie, Yuxin Wang, Zhendong Mao, and Yongdong Zhang.
\newblock Read like humans: Autonomous, bidirectional and iterative language modeling for scene text recognition.
\newblock In {\em Proceedings of the IEEE/CVF Conference on Computer Vision and Pattern Recognition}, pages 7098--7107, 2021.

\bibitem[\protect\citeauthoryear{Graves \bgroup \em et al.\egroup }{2006}]{ctc}
Alex Graves, Santiago Fern{\'a}ndez, Faustino Gomez, and J{\"u}rgen Schmidhuber.
\newblock Connectionist temporal classification: labelling unsegmented sequence data with recurrent neural networks.
\newblock In {\em Proceedings of the 23rd international conference on Machine learning}, pages 369--376, 2006.

\bibitem[\protect\citeauthoryear{Gupta \bgroup \em et al.\egroup }{2016}]{ST}
Ankush Gupta, Andrea Vedaldi, and Andrew Zisserman.
\newblock Synthetic data for text localisation in natural images.
\newblock In {\em Proceedings of the IEEE conference on computer vision and pattern recognition}, pages 2315--2324, 2016.

\bibitem[\protect\citeauthoryear{Jaderberg \bgroup \em et al.\egroup }{2014}]{MJ1}
Max Jaderberg, Karen Simonyan, Andrea Vedaldi, and Andrew Zisserman.
\newblock Synthetic data and artificial neural networks for natural scene text recognition.
\newblock {\em arXiv preprint arXiv:1406.2227}, 2014.

\bibitem[\protect\citeauthoryear{Jaderberg \bgroup \em et al.\egroup }{2016}]{MJ2}
Max Jaderberg, Karen Simonyan, Andrea Vedaldi, and Andrew Zisserman.
\newblock Reading text in the wild with convolutional neural networks.
\newblock {\em International journal of computer vision}, 116(1):1--20, 2016.

\bibitem[\protect\citeauthoryear{Jiang \bgroup \em et al.\egroup }{2023}]{union14m}
Qing Jiang, Jiapeng Wang, Dezhi Peng, Chongyu Liu, and Lianwen Jin.
\newblock Revisiting scene text recognition: A data perspective.
\newblock In {\em Proceedings of the IEEE/CVF International Conference on Computer Vision}, pages 20543--20554, 2023.

\bibitem[\protect\citeauthoryear{Karatzas \bgroup \em et al.\egroup }{2013}]{IC13}
Dimosthenis Karatzas, Faisal Shafait, Seiichi Uchida, Masakazu Iwamura, Lluis~Gomez i~Bigorda, Sergi~Robles Mestre, Joan Mas, David~Fernandez Mota, Jon~Almazan Almazan, and Lluis~Pere De~Las~Heras.
\newblock Icdar 2013 robust reading competition.
\newblock In {\em 2013 12th international conference on document analysis and recognition}, pages 1484--1493. IEEE, 2013.

\bibitem[\protect\citeauthoryear{Karatzas \bgroup \em et al.\egroup }{2015}]{IC15}
Dimosthenis Karatzas, Lluis Gomez-Bigorda, Anguelos Nicolaou, Suman Ghosh, Andrew Bagdanov, Masakazu Iwamura, Jiri Matas, Lukas Neumann, Vijay~Ramaseshan Chandrasekhar, Shijian Lu, et~al.
\newblock Icdar 2015 competition on robust reading.
\newblock In {\em 2015 13th international conference on document analysis and recognition (ICDAR)}, pages 1156--1160. IEEE, 2015.

\bibitem[\protect\citeauthoryear{Kingma and Ba}{2014}]{adam}
Diederik~P Kingma and Jimmy Ba.
\newblock Adam: A method for stochastic optimization.
\newblock {\em arXiv preprint arXiv:1412.6980}, 2014.

\bibitem[\protect\citeauthoryear{Mishra \bgroup \em et al.\egroup }{2012}]{IIIT}
Anand Mishra, Karteek Alahari, and CV~Jawahar.
\newblock Scene text recognition using higher order language priors.
\newblock In {\em BMVC-British machine vision conference}. BMVA, 2012.

\bibitem[\protect\citeauthoryear{Na \bgroup \em et al.\egroup }{2022}]{matrn}
Byeonghu Na, Yoonsik Kim, and Sungrae Park.
\newblock Multi-modal text recognition networks: Interactive enhancements between visual and semantic features.
\newblock In {\em European Conference on Computer Vision}, pages 446--463. Springer, 2022.

\bibitem[\protect\citeauthoryear{Phan \bgroup \em et al.\egroup }{2013}]{SVTP}
Trung~Quy Phan, Palaiahnakote Shivakumara, Shangxuan Tian, and Chew~Lim Tan.
\newblock Recognizing text with perspective distortion in natural scenes.
\newblock In {\em Proceedings of the IEEE International Conference on Computer Vision}, pages 569--576, 2013.

\bibitem[\protect\citeauthoryear{Qi and Su}{2017}]{ccloss}
Ce~Qi and Fei Su.
\newblock Contrastive-center loss for deep neural networks.
\newblock In {\em 2017 IEEE international conference on image processing (ICIP)}, pages 2851--2855. IEEE, 2017.

\bibitem[\protect\citeauthoryear{Risnumawan \bgroup \em et al.\egroup }{2014}]{CUTE}
Anhar Risnumawan, Palaiahankote Shivakumara, Chee~Seng Chan, and Chew~Lim Tan.
\newblock A robust arbitrary text detection system for natural scene images.
\newblock {\em Expert Systems with Applications}, 41(18):8027--8048, 2014.

\bibitem[\protect\citeauthoryear{Sheng \bgroup \em et al.\egroup }{2019}]{nrtr}
Fenfen Sheng, Zhineng Chen, and Bo~Xu.
\newblock Nrtr: A no-recurrence sequence-to-sequence model for scene text recognition.
\newblock In {\em 2019 International conference on document analysis and recognition (ICDAR)}, pages 781--786. IEEE, 2019.

\bibitem[\protect\citeauthoryear{Shi \bgroup \em et al.\egroup }{2016}]{CRNN}
Baoguang Shi, Xiang Bai, and Cong Yao.
\newblock An end-to-end trainable neural network for image-based sequence recognition and its application to scene text recognition.
\newblock {\em IEEE transactions on pattern analysis and machine intelligence}, 39(11):2298--2304, 2016.

\bibitem[\protect\citeauthoryear{Shi \bgroup \em et al.\egroup }{2018}]{aster}
Baoguang Shi, Mingkun Yang, Xinggang Wang, Pengyuan Lyu, Cong Yao, and Xiang Bai.
\newblock Aster: An attentional scene text recognizer with flexible rectification.
\newblock {\em IEEE transactions on pattern analysis and machine intelligence}, 41(9):2035--2048, 2018.

\bibitem[\protect\citeauthoryear{Sun \bgroup \em et al.\egroup }{2023}]{retentive}
Yutao Sun, Li~Dong, Shaohan Huang, Shuming Ma, Yuqing Xia, Jilong Xue, Jianyong Wang, and Furu Wei.
\newblock Retentive network: A successor to transformer for large language models.
\newblock {\em arXiv preprint arXiv:2307.08621}, 2023.

\bibitem[\protect\citeauthoryear{Van~der Maaten and Hinton}{2008}]{tsne}
Laurens Van~der Maaten and Geoffrey Hinton.
\newblock Visualizing data using t-sne.
\newblock {\em Journal of machine learning research}, 9(11), 2008.

\bibitem[\protect\citeauthoryear{Veit \bgroup \em et al.\egroup }{2016}]{cocotext}
Andreas Veit, Tomas Matera, Lukas Neumann, Jiri Matas, and Serge Belongie.
\newblock Coco-text: Dataset and benchmark for text detection and recognition in natural images.
\newblock {\em arXiv preprint arXiv:1601.07140}, 2016.

\bibitem[\protect\citeauthoryear{Wang \bgroup \em et al.\egroup }{2011}]{SVT}
Kai Wang, Boris Babenko, and Serge Belongie.
\newblock End-to-end scene text recognition.
\newblock In {\em 2011 International conference on computer vision}, pages 1457--1464. IEEE, 2011.

\bibitem[\protect\citeauthoryear{Wang \bgroup \em et al.\egroup }{2021}]{visionlan}
Yuxin Wang, Hongtao Xie, Shancheng Fang, Jing Wang, Shenggao Zhu, and Yongdong Zhang.
\newblock From two to one: A new scene text recognizer with visual language modeling network.
\newblock In {\em Proceedings of the IEEE/CVF International Conference on Computer Vision}, pages 14194--14203, 2021.

\bibitem[\protect\citeauthoryear{Wang \bgroup \em et al.\egroup }{2022}]{MGP}
Peng Wang, Cheng Da, and Cong Yao.
\newblock Multi-granularity prediction for scene text recognition.
\newblock In {\em European Conference on Computer Vision}, pages 339--355. Springer, 2022.

\bibitem[\protect\citeauthoryear{Wang \bgroup \em et al.\egroup }{2023}]{clipocr}
Zixiao Wang, Hongtao Xie, Yuxin Wang, Jianjun Xu, Boqiang Zhang, and Yongdong Zhang.
\newblock Symmetrical linguistic feature distillation with clip for scene text recognition.
\newblock In {\em Proceedings of the 31st ACM International Conference on Multimedia}, pages 509--518, 2023.

\bibitem[\protect\citeauthoryear{Xie \bgroup \em et al.\egroup }{2022}]{wordart}
Xudong Xie, Ling Fu, Zhifei Zhang, Zhaowen Wang, and Xiang Bai.
\newblock Toward understanding wordart: Corner-guided transformer for scene text recognition.
\newblock In {\em European Conference on Computer Vision}, pages 303--321. Springer, 2022.

\bibitem[\protect\citeauthoryear{Yang \bgroup \em et al.\egroup }{2022}]{DiG}
Mingkun Yang, Minghui Liao, Pu~Lu, Jing Wang, Shenggao Zhu, Hualin Luo, Qi~Tian, and Xiang Bai.
\newblock Reading and writing: Discriminative and generative modeling for self-supervised text recognition.
\newblock In {\em Proceedings of the 30th ACM International Conference on Multimedia}, pages 4214--4223, 2022.

\bibitem[\protect\citeauthoryear{Yu \bgroup \em et al.\egroup }{2020}]{SRN}
Deli Yu, Xuan Li, Chengquan Zhang, Tao Liu, Junyu Han, Jingtuo Liu, and Errui Ding.
\newblock Towards accurate scene text recognition with semantic reasoning networks.
\newblock In {\em Proceedings of the IEEE/CVF Conference on Computer Vision and Pattern Recognition}, pages 12113--12122, 2020.

\bibitem[\protect\citeauthoryear{Yue \bgroup \em et al.\egroup }{2020}]{robustscanner}
Xiaoyu Yue, Zhanghui Kuang, Chenhao Lin, Hongbin Sun, and Wayne Zhang.
\newblock Robustscanner: Dynamically enhancing positional clues for robust text recognition.
\newblock In {\em European Conference on Computer Vision}, pages 135--151. Springer, 2020.

\bibitem[\protect\citeauthoryear{Zhang \bgroup \em et al.\egroup }{2017}]{uber}
Ying Zhang, Lionel Gueguen, Ilya Zharkov, Peter Zhang, Keith Seifert, and Ben Kadlec.
\newblock Uber-text: A large-scale dataset for optical character recognition from street-level imagery.
\newblock In {\em SUNw: Scene Understanding Workshop-CVPR}, volume 2017, page~5, 2017.

\bibitem[\protect\citeauthoryear{Zhang \bgroup \em et al.\egroup }{2022}]{conclr}
Xinyun Zhang, Binwu Zhu, Xufeng Yao, Qi~Sun, Ruiyu Li, and Bei Yu.
\newblock Context-based contrastive learning for scene text recognition.
\newblock In {\em Proceedings of the AAAI Conference on Artificial Intelligence}, volume~36, pages 3353--3361, 2022.

\bibitem[\protect\citeauthoryear{Zhang \bgroup \em et al.\egroup }{2023}]{zhangbo}
Boqiang Zhang, Hongtao Xie, Yuxin Wang, Jianjun Xu, and Yongdong Zhang.
\newblock Linguistic more: Taking a further step toward efficient and accurate scene text recognition.
\newblock {\em arXiv preprint arXiv:2305.05140}, 2023.

\end{thebibliography}

\end{document}